\documentclass[10pt,conference]{IEEEtran}
\IEEEoverridecommandlockouts
\usepackage{xcolor}
\usepackage{cite}
\usepackage{amsmath,amssymb,amsfonts}
\usepackage{algorithmic}
\usepackage{graphicx}
\usepackage{textcomp}
\usepackage{xurl}
\usepackage[hidelinks]{hyperref}
\def\BibTeX{{\rm B\kern-.05em{\sc i\kern-.025em b}\kern-.08em
    T\kern-.1667em\lower.7ex\hbox{E}\kern-.125emX}}
 
\newcommand{\acceptedvenue}{Accepted for publication in the Proceedings of the 2027 IEEE Symposium Series on Computational Intelligence (IEEE SSCI 2027)} 
\begin{document}

\title{\vspace{-1.5cm}
    {\normalfont\normalsize\underline{\acceptedvenue}}\\[2ex]
    AV-FLARE: Feature-Level Reliability Estimation for Multimodal Emotion Recognition}

\author{
\IEEEauthorblockN{\textbf{Ines Jemmali}\textsuperscript{1,2}, \textbf{Alaa Nfissi}\textsuperscript{1,2}, \textbf{Wassim Bouachir}\textsuperscript{1} and \textbf{Nizar Bouguila}\textsuperscript{2}}
\IEEEauthorblockA{\textsuperscript{1}Université T\'ELUQ , Montr\'eal, Qu\'ebec, Canada \\
\textsuperscript{2}Concordia University, Montr\'eal, Qu\'ebec, Canada \\
 \{ines.jemmali, alaa.nfissi\}@mail.concordia.ca, wassim.bouachir@teluq.ca, nizar.bouguila@concordia.ca}

}

\maketitle

\begingroup
\renewcommand{\thefootnote}{}
\footnotetext{
© 2027 IEEE. Personal use of this material is permitted.  Permission from IEEE must be obtained for all other uses, in any current or future media, including reprinting/republishing this material for advertising or promotional purposes, creating new collective works, for resale or redistribution to servers or lists, or reuse of any copyrighted component of this work in other works.
}
\endgroup

\begin{abstract}
Multimodal emotion recognition (MMER) systems typically rely on learned feature representations and attention mechanisms that implicitly weight information across time and modalities. However, these mechanisms do not explicitly incorporate input reliability, despite the presence of noise in audio, occlusions in visual data, and the sparse and event-driven nature of affective cues. We propose AV-FLARE, a framework that explicitly estimates fine-grained reliability scores across both temporal and feature dimensions prior to temporal modeling and uses them to modulate feature representations. Reliability is computed using two complementary components: (i) a supervised network trained jointly with the classification objective, and (ii) an unsupervised Gaussian Mixture Model that models global feature distributions. These signals are combined to provide both task-driven and distribution-aware reliability estimates. To integrate modalities, we introduce a hierarchical cross-modal attention mechanism in which textual features are used to guide audio-visual fusion as a semantic reference modality. Experiments on the IEMOCAP dataset show that AV-FLARE improves over a no-reliability baseline, achieving 90.81\% accuracy and 81.69\% F1-score, corresponding to gains of +4.41\% and +8.56\%, respectively. These results suggest that incorporating explicit input reliability estimation may improve robustness in multimodal emotion recognition under the evaluated setting. Our source code is available at: \href{https://github.com/InesJemmali/Feature-Level-Reliability-Estimation-for-Multimodal-Emotion-Recognition.git}{https://github.com/InesJemmali/AV-FLARE}.
\end{abstract}

\begin{IEEEkeywords}
Multimodal Emotion Recognition, Deep Learning, Feature Reliability Estimation, Gaussian Mixture Model
\end{IEEEkeywords}

\section{Introduction}

Recognizing human emotions is a long-standing goal in affective computing~\cite{Filali2025ComparativeSO}. This task has practical applications in areas such as human-computer interaction and conversational systems, where recognizing a person's emotional state can help systems provide more appropriate responses. Multimodal emotion recognition (MMER) aims to identify emotional states by jointly analyzing information from multiple modalities, typically audio, video, and text~\cite{lian2023survey}. Each modality provides a different type of information. Audio captures acoustic characteristics such as tone, pitch, and speaking intensity. Video provides visual cues through facial expressions and movements, while text conveys the linguistic content of speech. By combining these complementary sources, MMER systems can capture emotional cues that may not be sufficiently represented in any single modality \cite{lian2023survey}. For example, a person's spoken words may not fully reflect their emotional state, while their tone of voice and facial expressions can provide additional information. While current systems achieve strong performance on benchmark datasets, generalization to real-world recordings with higher variability remains challenging.

A key challenge lies in the heterogeneity of signal quality across time and modalities. Acoustic inputs may contain noise and reverberation, which can obscure or distort speech features~\cite{cesarini2024reverb}. Non-speech segments such as silence are often uninformative for emotion recognition (ER) and are typically excluded or downweighted during processing~\cite{ma2019audio}. Visual streams are affected by illumination variations and motion-related distortions, which can reduce the reliability of extracted facial features~\cite{yan2002detection}. In addition, multimodal signals often exhibit uneven informativeness over time, where emotionally relevant cues are not uniformly distributed across an utterance~\cite{baltruvsaitis2018multimodal}.

Robustness in MMER is commonly addressed through adaptive fusion strategies or uncertainty estimation applied after temporal encoding. While these approaches improve decision-level stability, they do not explicitly regulate the influence of unreliable inputs during early representation learning, where noisy signals may already affect intermediate features. Standard self-attention performs similarity-based aggregation of input elements, enabling dynamic interaction modeling across sequences, but does not explicitly incorporate uncertainty or reliability into the feature aggregation process. This can make learned representations sensitive to noise and data-dependent attention patterns. This has motivated increasing interest in uncertainty-aware and confidence-based methods for multimodal learning, aiming to improve robustness under noisy or incomplete inputs~\cite{tellamekala2023cold, zhu2023dynamic}.

In this work, we propose AV-FLARE, a framework that introduces explicit reliability estimation prior to temporal modeling. The key idea is to assess the quality of audio-visual representations at an early stage and use these estimates to modulate feature representations before sequence modeling, thereby reducing the influence of unreliable inputs while preserving informative signals. We introduce a dual reliability estimation strategy combining supervised and unsupervised components. The supervised component learns task-aligned reliability signals jointly with the classification objective, while the unsupervised component models global feature distribution structure to identify atypical or low-density patterns. These complementary signals are fused into a unified reliability estimate used for feature modulation. We further apply a hierarchical cross-modal attention mechanism that integrates modalities, where textual information provides a guiding context for aligning audio and visual representations at a higher semantic level.

We evaluate AV-FLARE on the IEMOCAP dataset~\cite{busso2008iemocap}. The proposed method achieves 90.81\% accuracy and 81.69\% F1-score, outperforming a no-reliability baseline by +4.41\% and +8.56\%, respectively. Ablation studies show that both reliability components contribute consistently and provide complementary benefits.

Our contributions are summarized as follows:
\begin{itemize}
\item A framework for explicit reliability estimation at the feature level prior to temporal modeling in multimodal emotion recognition.
\item A dual strategy combining supervised and unsupervised signals for robust reliability modeling and feature modulation.
\item A hierarchical cross-modal fusion mechanism guided by textual context for improved multimodal alignment.
\end{itemize}

Code is available at  \href{https://github.com/InesJemmali/Feature-Level-Reliability-Estimation-for-Multimodal-Emotion-Recognition.git}{\url{https://github.com/InesJemmali/Feature-Level-Reliability-Estimation-for-Multimodal-Emotion-Recognition.git}}. 

\section{Related Work}
\label{sec:relatedwork}

Emotion recognition has evolved from unimodal signal processing to multimodal architectures that integrate speech, visual, and textual modalities to exploit their complementary information. Early speech-based approaches demonstrated that emotional patterns can be learned directly from raw audio signals using deep architectures such as CNN-GRU models, without relying on handcrafted features \cite{nfissi2024unlocking}. Subsequent work introduced learnable time-frequency representations combined with recurrent networks to improve robustness under noisy acoustic conditions \cite{nfissi2024learnable}, highlighting the sensitivity of speech-based models to real-world signal variability.

To enhance robustness, MMER systems incorporate visual and textual modalities alongside acoustic inputs. Convolutional architectures remain widely used for acoustic feature extraction due to their ability to capture local acoustic features \cite{nfissi2022cnn}. In the visual domain, architectures such as RepVGG provide efficient and effective feature extraction for facial representations \cite{ding2021repvgg, wei2022fv2es}. For text, pretrained language models such as BERT and ALBERT are commonly used to extract contextual semantic embeddings \cite{devlin2019bert, DBLP:conf/iclr/LanCGGSS20}. Fusion strategies have evolved from early concatenation and late fusion methods \cite{zhao22k_interspeech} to attention-based mechanisms enabling more flexible cross-modal interactions \cite{mocanu2023multimodal, Nfissi2025Multimodal}.

Despite these advances, many multimodal architectures implicitly assume uniform or uncalibrated reliability across time and modalities. While self-attention mechanisms enable dynamic weighting based on content similarity, they do not explicitly model signal quality or uncertainty. Consequently, noisy or low-quality inputs may still influence learned representations depending on data distribution and attention patterns, particularly in real-world scenarios with asynchronous or degraded modalities.

To address robustness, a growing body of work incorporates uncertainty or reliability estimation into emotion recognition systems. Probabilistic approaches such as Gaussian Mixture Regression model uncertainty in continuous affect prediction by capturing relationships between feature distributions and emotional states \cite{dang2017investigation}. MC Dropout \cite{gal2016dropout} and evidential learning \cite{sensoy2018evidential} estimate predictive uncertainty at the output level, without modifying intermediate representations. At the fusion stage, dynamic confidence-aware methods estimate modality-level reliability using learned networks and apply it as weighting during feature aggregation \cite{zhu2023dynamic}, while COLD Fusion introduces uncertainty-aware temporal modeling through calibrated latent representations \cite{tellamekala2023cold}.

Complementary to this, feature reweighting mechanisms such as Squeeze-and-Excitation blocks \cite{hu2018squeeze} and attention-based recalibration methods \cite{tao2024depmstat} adaptively modulate feature importance using global or content-based context statistics. In particular, Squeeze-and-Excitation networks perform channel-wise recalibration by modeling inter-channel dependencies through globally aggregated feature descriptors. Similarly, attention-based architectures compute feature weights based on learned similarity relationships between input elements. However, these methods rely on content-driven statistics rather than explicit reliability or uncertainty modeling, and they do not incorporate mechanisms to assess input quality prior to temporal representation learning.

Overall, existing methods improve robustness through uncertainty estimation, confidence-aware fusion, or feature reweighting. However, the reviewed approaches do not explicitly combine frame and channel reliability estimation using complementary supervised and unsupervised estimators to recalibrate audio and visual features before temporal representation learning. In contrast, the proposed method integrates these components to assess feature reliability and modulate representations before temporal modeling, providing a complementary approach to handling unreliable inputs in MMER.

\section{Proposed Method}
\label{sec:method}

\subsection{Architecture Overview}
\begin{figure*}[htbp]
\scriptsize
\centerline{\includegraphics[scale=0.1]{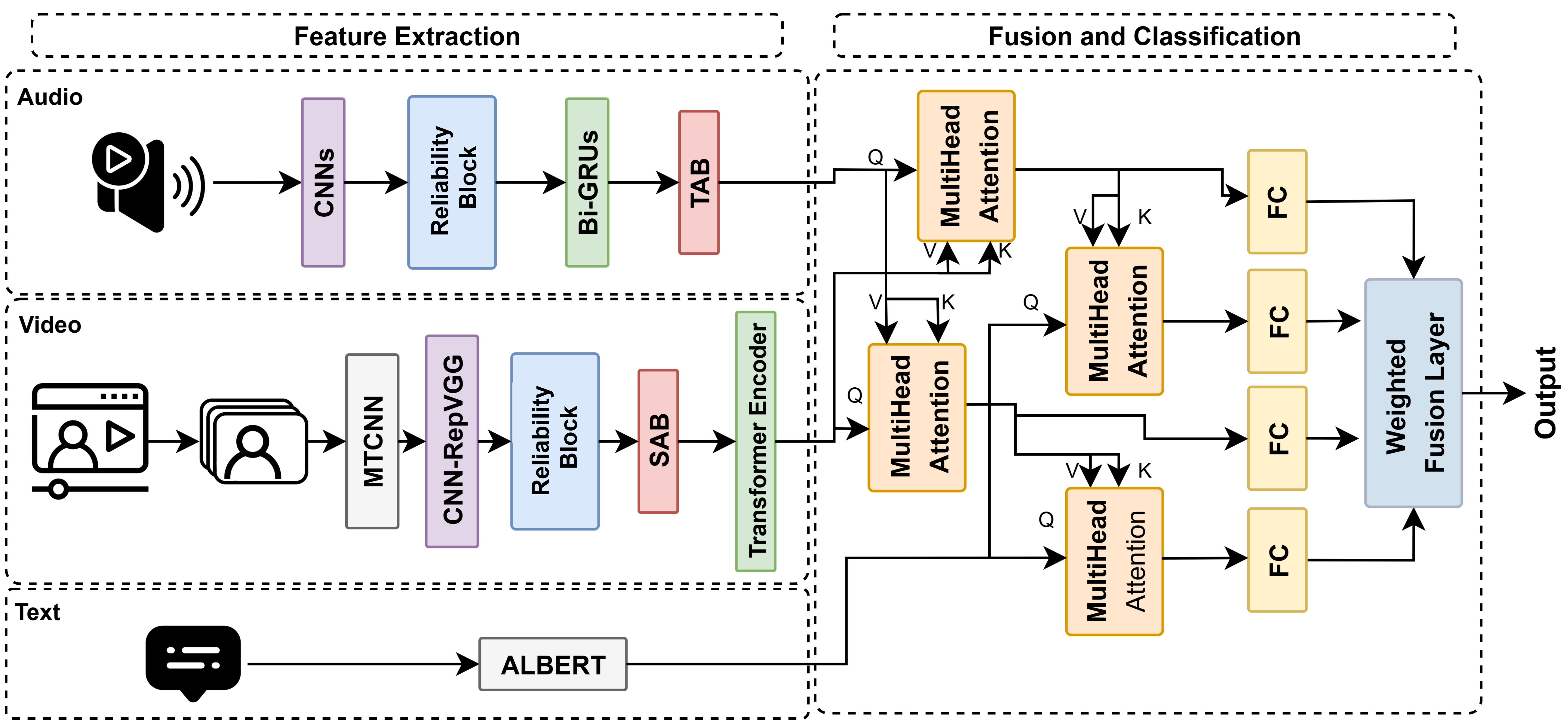}}
    \caption{
    Overview of the AV-FLARE architecture with audio-visual reliability estimation block and hierarchical cross-modal fusion.}
   
    \label{fig:archi}
\end{figure*}

AV-FLARE (Audio-Visual Feature-Level Reliability Estimation) addresses the challenge of unreliable and heterogeneous multimodal inputs by introducing explicit feature-level reliability estimation prior to temporal encoding. The core idea is to jointly model task-dependent and distributional reliability signals and use them to modulate audio-visual representations before sequence modeling, as illustrated in Fig.~\ref{fig:archi}. This is achieved through a dual reliability estimation module that combines a supervised feedforward network trained jointly with the classification objective \cite{moon2020confidence} and an unsupervised Gaussian Mixture Model that captures the density structure of feature representations and identifies low-probability regions associated with potential noise or corrupted inputs \cite{zhang2025magmm, qiao2023smart}. The two signals are fused to produce per-frame and per-channel reliability maps that are applied as multiplicative weights on the feature representations. Each modality is processed using a dedicated encoder. Audio is encoded using a seven-layer 1D CNN followed by a four-layer bidirectional GRU, while video is encoded using a CNN-RepVGG backbone followed by a four-layer Transformer encoder. Text is encoded using a pretrained ALBERT model. Reliability modulation is applied only to audio and visual streams, as these modalities exhibit continuous temporal structure and are more susceptible to frame-level noise and corruption. In contrast, although textual inputs are sequential, they are typically processed at a coarser temporal granularity, where tokens capture semantically aggregated information rather than fine-grained temporal variations. As a result, frame-level reliability estimation is less directly applicable to textual representations~\cite{baltruvsaitis2018multimodal}. After reliability-weighted encoding, audio and visual representations are first aligned using cross-modal attention, producing a joint audio-visual representation. This is subsequently refined using textual features through a second cross-attention stage, where text provides high-level semantic guidance. Finally, a learnable fusion layer combines the resulting cross-modal representations to produce the final emotion prediction logits.

\subsection{Feature Extraction}

\subsubsection{Audio Encoder}
Raw waveforms are used directly to preserve full temporal structure. A seven-layer 1D CNN extracts hierarchical acoustic representations, followed by a four-layer bidirectional GRU for sequential modeling. This class of CNN-RNN architectures has demonstrated strong performance for speech-based emotion recognition tasks \cite{nfissi2022cnn}. A Temporal Attention Block (TAB) \cite{tao2024depmstat} further refines temporal saliency over the recurrent outputs. Before temporal modeling, CNN features are passed to the reliability estimation module, which computes per-frame and per-channel reliability scores used to modulate feature activations.

\subsubsection{Video Encoder}
Video input consists of aligned face frames extracted at fixed temporal intervals using MTCNN \cite{zhang2016joint}, which provides robust face localization and alignment under variations in pose and illumination. Spatial features are extracted using a CNN-RepVGG backbone \cite{ding2021repvgg, wei2022fv2es}, chosen for its efficient single-branch inference and stable convolutional representations, making it well-suited for integration within multimodal pipelines. A Spatial Attention Block (SAB) \cite{tao2024depmstat} further refines intra-frame feature saliency. As in the audio branch, features are passed to the reliability module before temporal encoding to mitigate the effect of irrelevant features. A four-layer Transformer encoder then models long-range temporal dependencies \cite{vaswani2017attention, bertasius2021space}, and outputs are projected to the shared embedding space.

\subsubsection{Text Encoder}
Text is encoded using ALBERT \cite{DBLP:conf/iclr/LanCGGSS20}, which reduces parameter redundancy through factorized embeddings and cross-layer parameter sharing compared to BERT \cite{devlin2019bert}. This design enables efficient encoding while maintaining strong semantic representations, making it well-suited for multimodal settings where computational resources must be balanced across modalities. The sentence representation is obtained from the final-layer \texttt{[CLS]} token, denoted as $\mathbf{h}_{\text{CLS}} = \mathbf{H}_L[:, 0]$, where $\mathbf{H}_L$ denotes the final transformer hidden states. This representation serves as a global semantic embedding for cross-modal fusion.

\subsection{Audio-Visual Reliability Estimation and Modulation}

\begin{figure*}
\scriptsize
    \centering
    \includegraphics[scale=0.12]{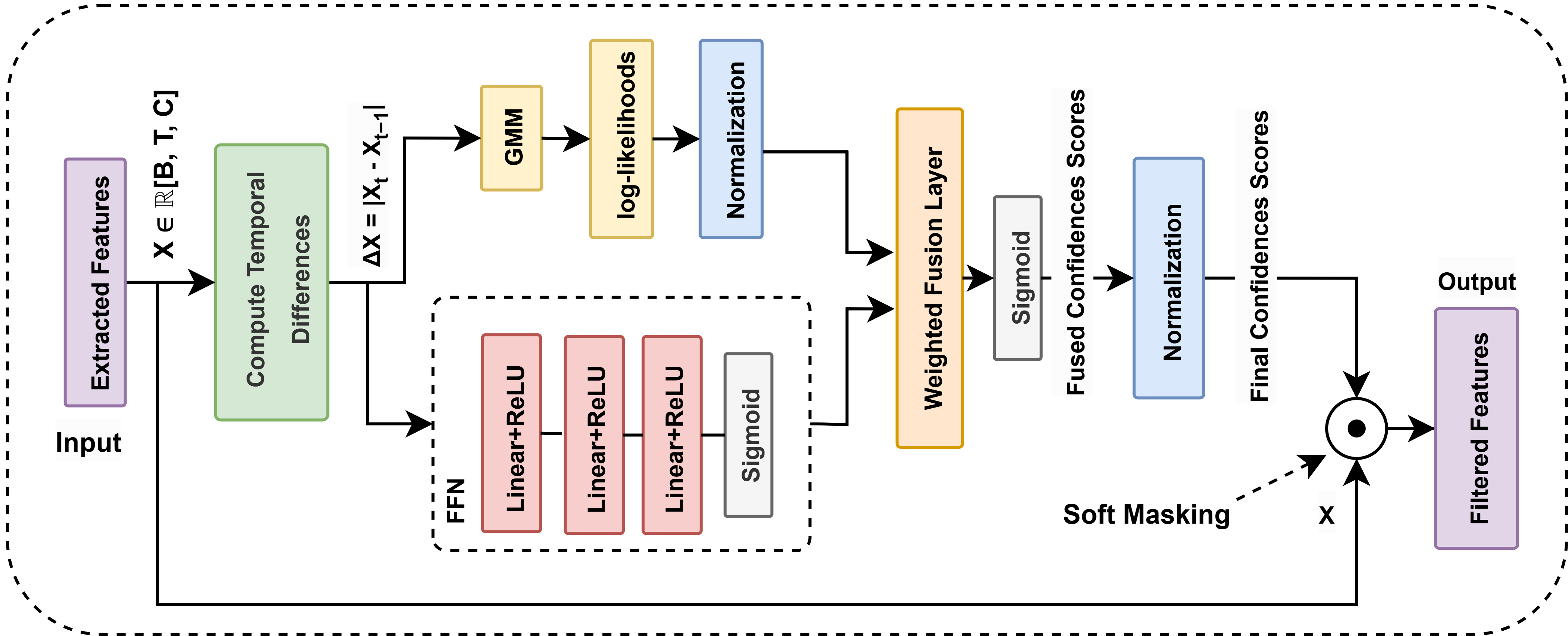}
    \caption{Architecture of the audio-visual reliability estimation block.}
    \label{fig:confidence_archi}
\end{figure*}

Fig.~\ref{fig:confidence_archi} illustrates the reliability estimation module, which operates on convolutional features before any recurrent or Transformer-based temporal modeling. Its role is to estimate per-frame and per-channel reliability scores and use them to modulate feature representations prior to sequence encoding. The module combines two complementary estimators: a supervised feedforward network (FFN) trained jointly with the classification objective, and an unsupervised Gaussian Mixture Model (GMM) that captures the global distribution of feature dynamics. Their outputs are fused to produce a unified reliability map applied via element-wise feature reweighting.

\paragraph{Supervised Reliability Estimation}
To capture task-dependent reliability, we estimate a discriminative reliability map from local temporal variations in the feature space. Given an input tensor $\mathbf{X} \in \mathbb{R}^{B \times T \times C}$, we compute frame-wise temporal differences as $\mathbf{D}_{b,t,c} = |\mathbf{X}_{b,t,c} - \mathbf{X}_{b,t-1,c}|$. These variations are then used to estimate supervised reliability through a lightweight feedforward network defined in Eq.~\ref{eq:rel_sup}.
\begin{equation}
\mathbf{R}_{\text{sup}} = \sigma(\text{MLP}(\mathbf{D}))
\label{eq:rel_sup}
\end{equation}
\noindent This design is motivated by prior work on confidence-aware deep learning. \cite{moon2020confidence} shows that jointly learning task-dependent reliability like confidence signals with the primary objective improves robustness to noisy or ambiguous inputs.

\paragraph{Unsupervised Reliability Estimation}

In parallel, we estimate distributional reliability using a GMM to capture the global structure of feature variations and identify atypical patterns in the feature space. Temporal differences are flattened as $\Delta\mathbf{X}_{\text{flat}} \in \mathbb{R}^{(B \cdot T) \times C}$. We model their distribution using a mixture density, as shown in Eq.~\ref{eq:gmm}. This formulation allows us to model data density explicitly, where low-probability regions correspond to potentially noisy or corrupted feature variations. Reliability is then derived from the log-likelihood of each frame's temporal difference vector, as given in Eq.~\ref{eq:gmm_logprob}. To ensure compatibility with the supervised signal and stable gating, the raw log-likelihoods are normalized to $[0, 1]$ using min-max normalization, as shown in Eq.~\ref{eq:gmm_norm}.
\begin{equation}
    p(\mathbf{x}) = \sum_{k=1}^{K} \pi_k \mathcal{N}(\mathbf{x};\, \mu_k, \Sigma_k)
    \label{eq:gmm}
\end{equation}
\begin{equation}
    \tilde{\mathcal{R}}_{t}^{\text{unsup}} = \log p(\Delta \mathbf{x}_t)
    \label{eq:gmm_logprob}
\end{equation}
\begin{equation}
    \mathcal{R}_{t}^{\text{unsup}} = 
    \frac{\tilde{\mathcal{R}}_{t}^{\text{unsup}} - \min_t \tilde{\mathcal{R}}_{t}^{\text{unsup}}}
         {\max_t \tilde{\mathcal{R}}_{t}^{\text{unsup}} - \min_t \tilde{\mathcal{R}}_{t}^{\text{unsup}} + \epsilon}
    \label{eq:gmm_norm}
\end{equation}
\noindent This provides a distribution-aware reliability signal that complements discriminative estimation by capturing feature atypicality, which has been shown to be effective in mixture-based uncertainty modeling for high-dimensional representations \cite{zhang2025magmm}. The model is re-estimated at each epoch to track the evolving feature distribution, following alternating
optimization strategies in representation learning where model parameters and latent assignments are updated iteratively \cite{caron2018deep}.

\paragraph{Reliability Fusion}
We combine the supervised and unsupervised reliability signals using a learnable gating mechanism that adapts their contributions based on input context. The gating weights and fused reliability are defined in Eqs.~\ref{eq:gating} and~\ref{eq:fusion}, respectively. This formulation provides an adaptive integration of heterogeneous reliability signals without imposing fixed weighting assumptions, allowing the model to adjust the contribution of each signal during training.
\begin{equation}
\mathbf{w} = \sigma(\text{Linear}([\mathbf{R}_{\text{sup}}, \mathbf{R}_{\text{unsup}}]))
\label{eq:gating}
\end{equation}
\begin{equation}
\mathbf{R}_{\text{fused}} = \mathbf{w} \cdot \mathbf{R}_{\text{sup}} + (1 - \mathbf{w}) \cdot \mathbf{R}_{\text{unsup}}
\label{eq:fusion}
\end{equation}

\subsection{Hierarchical Cross-Modal Attention}

Following temporal encoding, audio and visual representations are integrated using a hierarchical cross-modal attention mechanism that first aligns modalities at the feature level and then refines them using semantic guidance from text. Let $\mathbf{A} \in \mathbb{R}^{B \times T_a \times D}$ and $\mathbf{V} \in \mathbb{R}^{B \times T_v \times D}$ denote the temporally encoded audio and visual embeddings. We apply bidirectional cross-attention to achieve audio-visual alignment (Eq.~\ref{eq:av_align}).
\begin{equation}
\mathbf{V}' = \text{MHA}(\mathbf{V}, \mathbf{A}, \mathbf{A}), \quad
\mathbf{A}' = \text{MHA}(\mathbf{A}, \mathbf{V}, \mathbf{V})
\label{eq:av_align}
\end{equation}
\noindent The aligned representations are aggregated into fixed-length embeddings using masked mean pooling to obtain $\bar{\mathbf{A}}'$ and $\bar{\mathbf{V}}'$. To incorporate high-level semantics, text representation $\mathbf{t} \in \mathbb{R}^{B \times D_t}$ is used as the query over the aligned audio-visual features. This refinement step is defined in Eq.~\ref{eq:text_refine}:
\begin{equation}
\mathbf{z}_{ta} = \text{MHA}(\mathbf{t}, \bar{\mathbf{A}}', \bar{\mathbf{A}}'), \quad
\mathbf{z}_{tv} = \text{MHA}(\mathbf{t}, \bar{\mathbf{V}}', \bar{\mathbf{V}}')
\label{eq:text_refine}
\end{equation}
\noindent This hierarchical design enables low-level cross-modal alignment between audio and vision while leveraging text as a semantic anchor to refine the fused representation at a higher level of abstraction.

\subsection{Fusion and Classification}

The cross-modal representations are first projected into class-wise logits, yielding $\mathbf{z}_{va}, \mathbf{z}_{av}, \mathbf{z}_{ta}, \mathbf{z}_{tv} \in \mathbb{R}^{B \times C}$. To aggregate these predictions, we construct a modality-wise logit tensor and apply a learnable weighted fusion mechanism defined in Eq.~\ref{eq:logit_fusion}:
\begin{equation}
\mathbf{Z} \in \mathbb{R}^{B \times C \times 4}, \quad
\mathbf{z}_{\text{fused}} = \mathbf{Z} \cdot \mathbf{w}
\label{eq:logit_fusion}
\end{equation}
\noindent where $\mathbf{w}$ is a learnable weighting vector that assigns a global weight to each cross-modal prediction pathway. These weights are learned during training and remain fixed across input samples at inference. The resulting weighted sum performs late-stage decision fusion by combining the four prediction pathways according to their learned relative importance.

\subsection{Training Objective}

We optimize the model using a combination of focal loss for classification and two reliability alignment regularization terms that encourage consistency between predicted reliability scores, prediction errors, and predictive uncertainty. To address class imbalance in IEMOCAP, we adopt focal loss \cite{lin2017focal} as the main classification objective defined in Eq.~\ref{eq:focal}.
\begin{equation}
\mathcal{L}_{\text{focal}} = -\alpha_t (1 - p_t)^{\gamma} \log(p_t)
\label{eq:focal}
\end{equation}
\noindent where $p_t$ is the predicted probability of the true class, $\alpha_t$ is a class-weighting factor, and $\gamma = 2$ controls the down-weighting of easy examples. To encourage meaningful calibration of the reliability module, we introduce two reliability regularization terms. For each modality $m \in \{\text{audio}, \text{video}\}$, let $\mathbf{S}^{(m)}$ denote the reliability estimates produced by the reliability module, $\mathbf{E}$ the binary error indicator, and $\mathbf{H}$ the entropy-based predictive uncertainty. Specifically, $E_i=1$ if the prediction for sample $i$ is incorrect and $E_i=0$ if it is correct. The entropy $H_i=-\sum_c p_{i,c}\log(p_{i,c})$ measures the predictive uncertainty for sample $i$, where $p_{i,c}$ denotes the model's predicted probability for class $c$. The error alignment regularizer is defined in Eq.~\ref{eq:rel_align}, while the second term aligns reliability scores with predictive uncertainty, as defined in Eq.~\ref{eq:unc_align}.
\begin{equation}
\mathcal{R}_{\text{align}}^{(m)} = 1 + \rho\left(\mathbf{S}^{(m)}, \mathbf{E}\right)
\label{eq:rel_align}
\end{equation}
\begin{equation}
\mathcal{R}_{\text{unc}}^{(m)} = 1 + \rho\left(\mathbf{S}^{(m)}, \mathbf{H}\right)
\label{eq:unc_align}
\end{equation}
\noindent where $\rho(\cdot,\cdot)$ denotes Pearson correlation. Minimizing these terms encourages reliability scores to be negatively correlated with both prediction errors and predictive uncertainty. This regularization guides the model toward more meaningful confidence estimation by encouraging the reliability module to assign lower scores to samples linked to incorrect predictions or higher predictive uncertainty. However, these relationships are encouraged statistically across samples and are not explicitly enforced for each individual sample. The final objective joins the focal loss with both regularization terms:
\begin{equation}
\mathcal{L} = \mathcal{L}_{\text{focal}} + \frac{\beta}{|M|} \sum_{\substack{m \in M}}
\mathcal{R}_{\text{unc}}^{(m)} + \frac{\beta}{|M|} \sum_{\substack{m \in M}}
\mathcal{R}_{\text{align}}^{(m)}
\label{eq:final_obj}
\end{equation}
\noindent where $M =\{\text{audio}, \text{video}\}$ and $\beta = 0.2$ is selected on the validation set. The two regularization terms jointly encourage reliability scores to reflect both prediction correctness and model uncertainty.

\section{Experiments and Results}
\label{sec:results}

\subsection{Dataset}

We evaluate AV-FLARE on the IEMOCAP dataset \cite{busso2008iemocap}, a widely used benchmark for multimodal emotion recognition. It contains approximately 12 hours of dyadic audiovisual interactions with both scripted and improvised scenarios. Following standard practice \cite{jin2015speech}, we focus on four emotion classes: \textit{Happy} (merged with \textit{Excited}), \textit{Sad}, \textit{Angry}, and \textit{Neutral}. The dataset is randomly divided at the utterance level into training, validation, and test sets, using an 80/10/10 split, respectively.
\begin{table}[t]
\scriptsize
    \centering
    \caption{AV-FLARE performance on the IEMOCAP dataset (in \%)}

    \begin{tabular}{lcccc}
        \hline
        \textbf{Class} & \textbf{Accuracy} & \textbf{Precision} & \textbf{Recall} & \textbf{F1-score} \\
       \hline
        Angry & 91.91 & 83.33 & 74.07 & 78.43 \\
        Happy & 88.97 & 78.65 & 86.42 & 82.35 \\
        Neutral & 87.87 & 80.25 & 79.27 & 79.75 \\
        Sad & 94.49 & 87.04 & 85.45 & 86.24 \\
        \hline
        \textbf{Average} & \textbf{90.81} & \textbf{82.32} & \textbf{81.30} & \textbf{81.69} \\
        \hline
    \end{tabular}
    \label{tab:results}
\end{table}

\begin{figure}[t]
\scriptsize
\centering
    \includegraphics[scale=0.38]{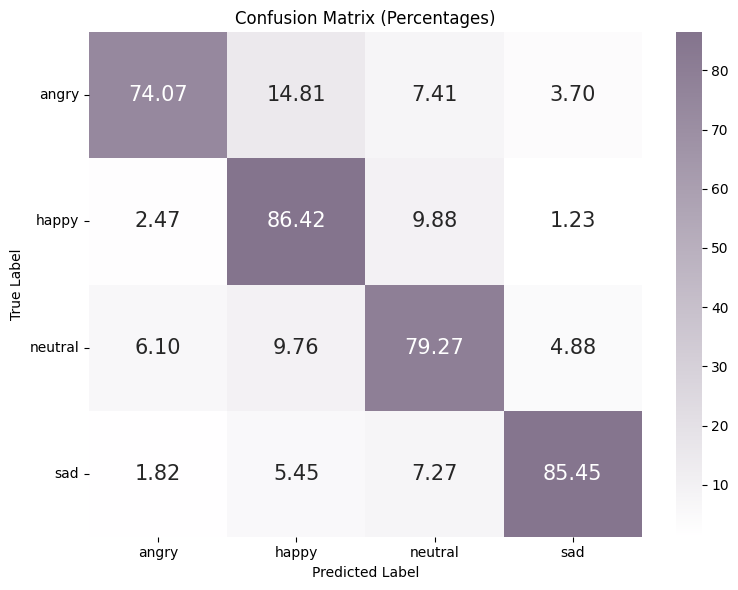}
    \caption{Confusion matrix on the IEMOCAP dataset.}
    \label{fig:confusionmatrix}
\end{figure}
\begin{table}[h]
\tiny
    \centering
    \caption{Comparison with recent methods on IEMOCAP dataset (in \%)}
    \resizebox{\linewidth}{!}{%
    \begin{tabular}{lcccc}
         \hline
         \textbf{Method} & \textbf{Accuracy} & \textbf{Precision} & \textbf{Recall} & 
         \textbf{F1-score} \\
     \hline
          M3ER \cite{mittal2020m3er} & 82.7 & NA & NA & 82.4 \\
           Liu et al. \cite{liu2024noise} & 84.3 & NA & NA & \textbf{84.5}\\  
          Fu et al. \cite{fu2024lmr} & 82.73 & NA & NA & 81.48 \\
           Shayaninasab et al. \cite{shayaninasab2024multi}& 75.42 & 76.1 & 75.42 & 75.35 \\
           Hosseini et al. \cite{hosseini2024multimodal} & 82.9 & NA & NA & NA \\
          Majumder et al. \cite{majumder2018multimodal} & 76.5 & NA & NA & 76.8 \\
           Nfissi et al. \cite{Nfissi2025Multimodal} & 88.31 & 76.53 &76.47 & 76.43 \\
        \hline
         \textbf{AV-FLARE (Ours)} & \textbf{90.81} & \textbf{82.32} & \textbf{81.30} & 81.69 \\
      \hline
    \end{tabular}}
    \label{tab:sota_comparison}
\end{table}
\begin{table*}[t]
\scriptsize
\caption{Ablation Study Evaluating the Impact of Reliability Estimation Components (FFN and GMM) on IEMOCAP (in \%)}
\begin{center}
\begin{tabular}{|l|cc|cc|cc|cc|cc|}
\hline
\textbf{Methods} & \multicolumn{2}{c|}{\textbf{Angry}} & \multicolumn{2}{c|}{\textbf{Happy}} & \multicolumn{2}{c|}{\textbf{Neutral}} & \multicolumn{2}{c|}{\textbf{Sad}} & \multicolumn{2}{c|}{\textbf{Average}} \\
\cline{2-11}
& Acc & F1 & Acc & F1 & Acc & F1 & Acc & F1 & Acc & F1 \\
\hline
No-Rel & 89.71 & 75.44 & 83.46 & 73.68 & 81.99 & 68.39 & 90.44 & 75.00 & 86.40 & 73.13 \\
GMM-Only & 87.87 & 70.27 & 86.03 & 76.54 & 81.62 & 69.14 & 93.75 & 84.40 & 87.32 & 75.09 \\
FFN-Only & 90.44 & 76.36 & 88.24 & 79.22 & 84.19 & 74.85 & 90.80 & 77.06 & 88.42 & 76.88 \\
\hline
\textbf{AV-FLARE (Full)} & \textbf{91.91} & \textbf{78.43} & \textbf{88.97} & \textbf{82.35} & \textbf{87.87} & \textbf{79.75} & \textbf{94.49} & \textbf{86.24} & \textbf{90.81} & \textbf{81.69} \\
\hline
\end{tabular}
\label{tab:ablationstudy}
\end{center}
\end{table*} 

\subsection{Experimental Setup}

The model is trained for 20 epochs using Adam with batch size 16. Learning rates are set to $4\mathrm{e}{-5}$ for multimodal modules and $4\mathrm{e}{-7}$ for ALBERT to stabilize fine-tuning.

All hyperparameters are selected on the validation set and fixed across experiments. Audio signals are peak-normalized, and face frames are detected using MTCNN. For reliability modeling, the number of GMM components is selected empirically, resulting in $K=3$ for audio and $K=4$ for video, balancing stability and representation of temporal variability.

\subsection{Results and Comparison with Recent Methods}

\subsubsection{Evaluation Metrics}

We report accuracy, precision, recall, and F1-score. These metrics jointly evaluate classification performance and class-wise balance, which is important under the inherent imbalance of emotional categories.

\subsubsection{Performance Analysis}

Table~\ref{tab:results} shows that AV-FLARE achieves 90.81\% accuracy and 81.69\% F1-score, indicating strong and balanced performance across all emotion categories. The results suggest that modeling reliability before temporal encoding improves robustness under heterogeneous multimodal conditions.

Performance is consistently higher for \textit{Sad} and \textit{Happy}, which typically exhibit more discriminative acoustic patterns such as prosodic variation and stable visual cues. These classes benefit most from reliable frame selection, since emotional expression is more temporally consistent. In contrast, \textit{Neutral} remains the most challenging class due to its low emotional salience and high overlap with other categories. This leads to ambiguous feature representations across modalities. \textit{Angry} shows lower recall, suggesting that some high-arousal but acoustically similar expressions are partially confused with \textit{Happy}, which is consistent with prior observations. 

The confusion matrix in Fig.~\ref{fig:confusionmatrix} further confirms these observations. The strong diagonal structure indicates stable predictions across all classes. Off-diagonal errors are mostly concentrated between emotionally similar categories, particularly between \textit{Angry} and \textit{Happy}, reflecting intrinsic affective ambiguity rather than model instability.

Overall, these results indicate that AV-FLARE maintains reliable performance even under noisy and overlapping emotional conditions, which is consistent with the proposed feature-level reliability modeling.

\subsubsection{Comparison with Recent Methods}

Table~\ref{tab:sota_comparison} demonstrates that AV-FLARE achieves the highest accuracy among recent MMER approaches on IEMOCAP, outperforming methods such as \cite{Nfissi2025Multimodal,liu2024noise,hosseini2024multimodal}. The improvement in accuracy indicates more consistent global predictions, while maintaining competitive F1-score. A key observation is that several recent approaches achieve competitive F1-scores but often at the cost of lower accuracy or incomplete metric reporting. This makes direct comparison challenging in some cases. In contrast, AV-FLARE maintains a balanced trade-off between precision and recall while improving overall classification accuracy.

The consistent gains suggest that explicitly modeling feature reliability at early stages reduces the impact of noisy or inconsistent multimodal inputs before temporal modeling and fusion. This leads to more stable representations and improved generalization across emotional categories.

\section{Ablation Study}

We evaluate the contribution of reliability estimation using four configurations: without reliability (No-Rel), GMM-only, FFN-only, and the full AV-FLARE model. 

As shown in Table~\ref{tab:ablationstudy}, both FFN and GMM separately improve performance over the baseline, showing that reliability modeling helps representation learning. FFN provides stronger gains by capturing task-specific temporal variations, while GMM improves robustness through global distribution modeling. The full model achieves the best results with 90.81\% accuracy and 81.69\% F1-score, outperforming the baseline by +4.41\% and +8.56\% respectively. Improvements are consistent across all emotion classes, with larger gains on \textit{Neutral} and \textit{Sad}, where noise and ambiguity are more pronounced. These results indicate that FFN and GMM are complementary: FFN captures local temporal reliability, while GMM provides global statistical regularization. Their combination leads to more stable and noise-robust multimodal representations.

\section{Conclusion}
\label{sec:conclusion}
We presented AV-FLARE, a multimodal emotion recognition framework that models modality reliability using complementary supervised (FFN) and unsupervised (GMM) estimators. Per-frame, per-channel reliability scores are used to modulate audio-visual features before temporal modeling, reducing the impact of noisy or uninformative segments. Cross-modal attention integrates complementary signals, while a trainable fusion mechanism adaptively combines modalities for final prediction. Experiments on IEMOCAP demonstrate that AV-FLARE achieves 90.81\% accuracy and 81.69\% F1-score, consistently outperforming ablation baselines. The results confirm that early reliability modeling improves robustness in multimodal emotion recognition. Some confusion remains between closely related emotions such as \textit{Angry} and \textit{Happy}, which reflects intrinsic ambiguity in affective expressions rather than model limitation. Future work will investigate generalization to additional datasets such as MSP-IMPROV and MELD, as well as robustness under controlled noise conditions. We also plan to explore class-aware reliability modeling and contrastive multimodal learning to further improve discriminative capability across subtle emotional states.

\section*{Acknowledgment}
This work was supported by the Canada Research Chair in Artificial Intelligence for Suicide Prevention (CRC-2023-00036).

\bibliographystyle{IEEEtran}
\bibliography{references}

@article{busso2008iemocap,
  title={IEMOCAP: Interactive emotional dyadic motion capture database},
  author={Busso, Carlos and Bulut, Murtaza and Lee, Chi-Chun and Kazemzadeh, Abe and Mower, Emily and Kim, Samuel and Chang, Jeannette N and Lee, Sungbok and Narayanan, Shrikanth S},
  journal={Language resources and evaluation},
  volume={42},
  number={4},
  pages={335--359},
  year={2008},
  publisher={Springer}
}

@inproceedings{moon2020confidence,
  title={Confidence-aware learning for deep neural networks},
  author={Moon, Jooyoung and Kim, Jihyo and Shin, Younghak and Hwang, Sangheum},
  booktitle={international conference on machine learning},
  pages={7034--7044},
  year={2020},
  organization={PMLR}
}

@inproceedings{hu2018squeeze,
  title={Squeeze-and-excitation networks},
  author={Hu, Jie and Shen, Li and Sun, Gang},
  booktitle={Proceedings of the IEEE conference on computer vision and pattern recognition},
  pages={7132--7141},
  year={2018}
}

@article{zhang2025magmm,
  title={MAGMM: A high-dimensional outlier detection algorithm based on a memory-augmented autoencoder and the Gaussian mixture model},
  author={Zhang, Zhongping and Wang, Kuo and Wang, Zhongman and Li, Junji},
  journal={Information Sciences},
  pages={122574},
  year={2025},
  publisher={Elsevier}
}

@article{qiao2023smart,
  title={Smart grid outlier detection based on the minorization--maximization algorithm},
  author={Qiao, Lina and Gao, Wengen and Li, Yunfei and Guo, Xinxin and Hu, Pengfei and Hua, Feng},
  journal={Sensors},
  volume={23},
  number={19},
  pages={8053},
  year={2023},
  publisher={MDPI}
}

@inproceedings{caron2018deep,
  title={Deep clustering for unsupervised learning of visual features},
  author={Caron, Mathilde and Bojanowski, Piotr and Joulin, Armand and Douze, Matthijs},
  booktitle={Proceedings of the European conference on computer vision (ECCV)},
  pages={132--149},
  year={2018}
}

@article{ma2019audio,
  title={Audio-visual emotion fusion (AVEF): A deep efficient weighted approach},
  author={Ma, Yaxiong and Hao, Yixue and Chen, Min and Chen, Jincai and Lu, Ping and Ko{\v{s}}ir, Andrej},
  journal={Information Fusion},
  volume={46},
  pages={184--192},
  year={2019},
  publisher={Elsevier}
}

@article{sensoy2018evidential,
  title={Evidential deep learning to quantify classification uncertainty},
  author={Sensoy, Murat and Kaplan, Lance and Kandemir, Melih},
  journal={Advances in neural information processing systems},
  volume={31},
  year={2018}
}

@inproceedings{gal2016dropout,
  title={Dropout as a bayesian approximation: Representing model uncertainty in deep learning},
  author={Gal, Yarin and Ghahramani, Zoubin},
  booktitle={international conference on machine learning},
  pages={1050--1059},
  year={2016},
  organization={PMLR}
}

@inproceedings{yan2002detection,
  title={Detection and removal of lighting \& shaking artifacts in home videos},
  author={Yan, Wei-Qi and Kankanhalli, Mohan S},
  booktitle={Proceedings of the tenth ACM international conference on Multimedia},
  pages={107--116},
  year={2002}
}

@article{cesarini2024reverb,
  title={Reverb and noise as real-world effects in speech recognition models: a study and a proposal of a feature set},
  author={Cesarini, Valerio and Costantini, Giovanni},
  journal={Applied Sciences},
  volume={14},
  number={23},
  pages={11446},
  year={2024},
  publisher={MDPI}
}

@article{baltruvsaitis2018multimodal,
  title={Multimodal machine learning: A survey and taxonomy},
  author={Baltru{\v{s}}aitis, Tadas and Ahuja, Chaitanya and Morency, Louis-Philippe},
  journal={IEEE transactions on pattern analysis and machine intelligence},
  volume={41},
  number={2},
  pages={423--443},
  year={2018},
  publisher={IEEE}
}

@inproceedings{DBLP:conf/iclr/LanCGGSS20,
  author       = {Zhenzhong Lan and
                  Mingda Chen and
                  Sebastian Goodman and
                  Kevin Gimpel and
                  Piyush Sharma and
                  Radu Soricut},
  title        = {{ALBERT:} {A} Lite {BERT} for Self-supervised Learning of Language
                  Representations},
  booktitle    = {8th International Conference on Learning Representations, {ICLR} 2020,
                  Addis Ababa, Ethiopia, April 26-30, 2020},
  publisher    = {OpenReview.net},

 address   = {Addis Ababa, Ethiopia},

  year         = {2020},
  url          = {https://openreview.net/forum?id=H1eA7AEtvS},
  bibsource    = {dblp computer science bibliography, https://dblp.org}
}

@inproceedings{devlin2019bert,
  title={Bert: Pre-training of deep bidirectional transformers for language understanding},
  author={Devlin, Jacob and Chang, Ming-Wei and Lee, Kenton and Toutanova, Kristina},
  booktitle={Proceedings of the 2019 conference of the North American chapter of the association for computational linguistics: human language technologies, volume 1 (long and short papers)},
  pages={4171--4186},
  year={2019}
}

@article{zhu2023dynamic,
  title={Dynamic confidence-aware multi-modal emotion recognition},
  author={Zhu, Qi and Zheng, Chuhang and Zhang, Zheng and Shao, Wei and Zhang, Daoqiang},
  journal={IEEE Transactions on Affective Computing},
  volume={15},
  number={3},
  pages={1358--1370},
  year={2023},
  publisher={IEEE}
}

@inproceedings{lin2017focal,
  title={Focal loss for dense object detection},
  author={Lin, Tsung-Yi and Goyal, Priya and Girshick, Ross and He, Kaiming and Doll{\'a}r, Piotr},
  booktitle={Proceedings of the IEEE international conference on computer vision},
  pages={2980--2988},
  year={2017}
}

@inproceedings{zhao22k_interspeech,
  title     = {{Multi-level Fusion of Wav2vec 2.0 and BERT for Multimodal Emotion Recognition}},
  author    = {{Zihan Zhao and Yanfeng Wang and Yu Wang}},
  year      = {{2022}},
  booktitle = {{Interspeech 2022}},
  pages     = {{4725--4729}},
  doi       = {{10.21437/Interspeech.2022-10230}},
  issn      = {{2958-1796}},
}

@article{Nfissi2025Multimodal,
	author = {Nfissi, Alaa and Jemmali, Ines and Bouachir, Wassim and Bouguila, Nizar},
	journal = {Proceedings of the Canadian Conference on Artificial Intelligence},
	year = {2025},
	month = {may 19},
	note = {https://caiac.pubpub.org/pub/433f13bv},
	publisher = {Canadian Artificial Intelligence Association (CAIAC)},
	title = {Multimodal {Emotion} {Recognition} with {Cross}-{Attentive} {Learning} and {Feature} {Fusion}},
}

@article{lian2023survey,
  title={A survey of deep learning-based multimodal emotion recognition: Speech, text, and face},
  author={Lian, Hailun and Lu, Cheng and Li, Sunan and Zhao, Yan and Tang, Chuangao and Zong, Yuan},
  journal={Entropy},
  volume={25},
  number={10},
  pages={1440},
  year={2023},
  publisher={MDPI}
}

@article{liu2024noise,
  title={Noise-resistant multimodal transformer for emotion recognition},
  author={Liu, Yuanyuan and Zhang, Haoyu and Zhan, Yibing and Chen, Zijing and Yin, Guanghao and Wei, Lin and Chen, Zhe},
  journal={International Journal of Computer Vision},
  pages={1--21},
  year={2024},
  publisher={Springer}
}

@article{zhang2016joint,
  title={Joint face detection and alignment using multitask cascaded convolutional networks},
  author={Zhang, Kaipeng and Zhang, Zhanpeng and Li, Zhifeng and Qiao, Yu},
  journal={IEEE signal processing letters},
  volume={23},
  number={10},
  pages={1499--1503},
  year={2016},
  publisher={IEEE}
}

@inproceedings{nfissi2022cnn,
  title={CNN-n-GRU: end-to-end speech emotion recognition from raw waveform signal using CNNs and gated recurrent unit networks},
  author={Nfissi, Alaa and Bouachir, Wassim and Bouguila, Nizar and Mishara, Brian L},
  booktitle={2022 21st IEEE International Conference on Machine Learning and Applications (ICMLA)},
  pages={699--702},
  year={2022},
  organization={IEEE}
}

@article{hosseini2024multimodal,
  title={Multimodal modelling of human emotion using sound, image and text fusion},
  author={Hosseini, Seyed Sadegh and Yamaghani, Mohammad Reza and Poorzaker Arabani, Soodabeh},
  journal={Signal, image and video processing},
  volume={18},
  number={1},
  pages={71--79},
  year={2024},
  publisher={Springer}
}

@article{vaswani2017attention,
  title={Attention is all you need},
  author={Vaswani, Ashish and Shazeer, Noam and Parmar, Niki and Uszkoreit, Jakob and Jones, Llion and Gomez, Aidan N and Kaiser, {\L}ukasz and Polosukhin, Illia},
  journal={Advances in neural information processing systems},
  volume={30},
  year={2017}
}

@inproceedings{bertasius2021space,
  title={Is space-time attention all you need for video understanding?},
  author={Bertasius, Gedas and Wang, Heng and Torresani, Lorenzo},
  booktitle={ICML},
  volume={2},
  number={3},
  pages={4},
  year={2021}
}

@inproceedings{ding2021repvgg,
  title={Repvgg: Making vgg-style convnets great again},
  author={Ding, Xiaohan and Zhang, Xiangyu and Ma, Ningning and Han, Jungong and Ding, Guiguang and Sun, Jian},
  booktitle={Proceedings of the IEEE/CVF conference on computer vision and pattern recognition},
  pages={13733--13742},
  year={2021}
}

@article{fu2024lmr,
  title={LMR-CBT: Learning modality-fused representations with CB-transformer for multimodal emotion recognition from unaligned multimodal sequences},
  author={Fu, Ziwang and Liu, Feng and Xu, Qing and Fu, Xiangling and Qi, Jiayin},
  journal={Frontiers of Computer Science},
  volume={18},
  number={4},
  pages={184314},
  year={2024},
  publisher={Springer}
}

@inproceedings{jin2015speech,
  title={Speech emotion recognition with acoustic and lexical features},
  author={Jin, Qin and Li, Chengxin and Chen, Shizhe and Wu, Huimin},
  booktitle={2015 IEEE international conference on acoustics, speech and signal processing (ICASSP)},
  pages={4749--4753},
  year={2015},
  organization={IEEE}
}

@article{tao2024depmstat,
  title={DepMSTAT: Multimodal spatio-temporal attentional transformer for depression detection},
  author={Tao, Yongfeng and Yang, Minqiang and Li, Huiru and Wu, Yushan and Hu, Bin},
  journal={IEEE Transactions on Knowledge and Data Engineering},
  volume={36},
  number={7},
  pages={2956--2966},
  year={2024},
  publisher={IEEE}
}

@article{majumder2018multimodal,
  title={Multimodal sentiment analysis using hierarchical fusion with context modeling},
  author={Majumder, Navonil and Hazarika, Devamanyu and Gelbukh, Alexander and Cambria, Erik and Poria, Soujanya},
  journal={Knowledge-based systems},
  volume={161},
  pages={124--133},
  year={2018},
  publisher={Elsevier}
}

@article{shayaninasab2024multi,
  title={Multi-Modal Emotion Recognition by Text, Speech and Video Using Pretrained Transformers},
  author={Shayaninasab, Minoo and Babaali, Bagher},
  journal={arXiv preprint arXiv:2402.07327},
  year={2024}
}

@inproceedings{mittal2020m3er,
  title={M3er: Multiplicative multimodal emotion recognition using facial, textual, and speech cues},
  author={Mittal, Trisha and Bhattacharya, Uttaran and Chandra, Rohan and Bera, Aniket and Manocha, Dinesh},
  booktitle={Proceedings of the AAAI conference on artificial intelligence},
  volume={34},
  number={02},
  pages={1359--1367},
  year={2020}
}

@article{wei2022fv2es,
  title={FV2ES: A fully end2end multimodal system for fast yet effective video emotion recognition inference},
  author={Wei, Qinglan and Huang, Xuling and Zhang, Yuan},
  journal={IEEE Transactions on Broadcasting},
  volume={69},
  number={1},
  pages={10--20},
  year={2022},
  publisher={IEEE}
}

@inproceedings{nfissi2024unlocking,
  title={Unlocking the emotional states of high-risk suicide callers through speech analysis},
  author={Nfissi, Alaa and Bouachir, Wassim and Bouguila, Nizar and Mishara, Brian},
  booktitle={2024 IEEE 18th International Conference on Semantic Computing (ICSC)},
  pages={33--40},
  year={2024},
  organization={IEEE}
}

@inproceedings{nfissi2024learnable,
  title={Learnable Deep Wavelet Packet Transform for Speech Emotion Recognition in High-Risk Suicide Calls},
  author={Nfissi, Alaa and Bouachir, Wassim and Bouguila, Nizar and Mishara, Brian},
  booktitle={2024 International Conference on Machine Learning and Applications (ICMLA)},
  pages={214--221},
  year={2024},
  organization={IEEE}
}

@article{mocanu2023multimodal,
  title={Multimodal emotion recognition using cross modal audio-video fusion with attention and deep metric learning},
  author={Mocanu, Bogdan and Tapu, Ruxandra and Zaharia, Titus},
  journal={Image and Vision Computing},
  volume={133},
  pages={104676},
  year={2023},
  publisher={Elsevier}
}

@inproceedings{dang2017investigation,
  title={An Investigation of Emotion Prediction Uncertainty Using Gaussian Mixture Regression.},
  author={Dang, Ting and Sethu, Vidhyasaharan and Epps, Julien and Ambikairajah, Eliathamby},
  booktitle={INTERSPEECH},
  pages={1248--1252},
  year={2017}
}

@article{Filali2025ComparativeSO,
  title={Comparative Study of Unimodal and Multimodal Systems Based on MNN},
  author={Hajar Filali and Chafik Boulealam and Hamid Tairi and Khalid El Fazazy and Adnane Mohamed Mahraz and Jamal Riffi},
  journal={Journal of Information Systems Engineering and Management},
  year={2025},

}

@article{tellamekala2023cold,
  title={COLD fusion: Calibrated and ordinal latent distribution fusion for uncertainty-aware multimodal emotion recognition},
  author={Tellamekala, Mani Kumar and Amiriparian, Shahin and Schuller, Bj{\"o}rn W and Andr{\'e}, Elisabeth and Giesbrecht, Timo and Valstar, Michel},
  journal={IEEE Transactions on Pattern Analysis and Machine Intelligence},
  volume={46},
  number={2},
  pages={805--822},
  year={2023},
  publisher={IEEE}
}

\end{document}